\documentclass[conference]{IEEEtran}
\IEEEoverridecommandlockouts
\usepackage{cite}
\usepackage{amsmath,amssymb,amsfonts}
\usepackage{algorithmic}
\usepackage{graphicx}
\usepackage{textcomp}
\usepackage{xcolor}
\usepackage[hidelinks]{hyperref}

\usepackage{booktabs}
\usepackage{multirow}
\usepackage{orcidlink}
\usepackage[capitalise]{cleveref}
\usepackage{threeparttable}

\def\BibTeX{{\rm B\kern-.05em{\sc i\kern-.025em b}\kern-.08em
    T\kern-.1667em\lower.7ex\hbox{E}\kern-.125emX}}
\begin{document}

\title{
    CNN-JEPA: Self-Supervised Pretraining Convolutional Neural Networks Using Joint Embedding Predictive Architecture

}


\author{
\IEEEauthorblockN{András Kalapos\,\orcidlink{0000-0002-9018-1372}, Bálint Gyires-Tóth\,\orcidlink{0000-0003-1059-9822}}
\IEEEauthorblockA{\textit{Department of Telecommunications and Artificial Intelligence, Faculty of Electrical Engineering and Informatics} \\
\textit{Budapest University of Technology and Economics}\\
Műegyetem rkp. 3., H-1111 Budapest, Hungary \\
\href{mailto:kalapos.andras@tmit.bme.hu}{kalapos.andras@tmit.bme.hu}
}
}

\maketitle


\begin{abstract}
    Self-supervised learning (SSL) has become an important approach in pretraining large neural networks, enabling unprecedented scaling of model and dataset sizes. While recent advances like I-JEPA have shown promising results for Vision Transformers, adapting such methods to Convolutional Neural Networks (CNNs) presents unique challenges. In this paper, we introduce CNN-JEPA, a novel SSL method that successfully applies the joint embedding predictive architecture approach to CNNs. Our method incorporates a sparse CNN encoder to handle masked inputs, a fully convolutional predictor using depthwise separable convolutions, and an improved masking strategy. We demonstrate that CNN-JEPA outperforms I-JEPA with ViT architectures on ImageNet-100, achieving a 73.3\% linear top-1 accuracy using a standard ResNet-50 encoder. Compared to other CNN-based SSL methods, CNN-JEPA requires 17-35\% less training time for the same number of epochs and approaches the linear and k-NN top-1 accuracies of BYOL, SimCLR, and VICReg.
    Our approach offers a simpler, more efficient alternative to existing SSL methods for CNNs, requiring minimal augmentations and no separate projector network. 
\end{abstract}


\section{Introduction}

Self-supervised learning (SSL) became an important approach in pretraining large neural networks, allowing model and dataset sizes to scale to levels that were previously unattainable. 
In computer vision, designing good self-supervised pretext learning tasks is a challenging problem, with many methods proposed in the literature.
One such approach is I-JEPA~\cite{SelfSupervisedLearningImages2023a} (Image Joint Embedding Predictive Architecture), which learns by predicting the latent representation of the masked patches from latent representations of provided context patches. This approach is particularly well-suited for Vision Transformers. 
However, adapting I-JEPA to Convolutional Neural Networks is non-trivial, as neither masking nor predicting from a dense feature map is straightforward. Computing feature maps for a masked input requires masked convolution and careful construction of the mask, that takes the network's downsampling into account. Predicting masked patches is also challenging because it requires a sufficiently large receptive field capable of predicting the latent representation based on the provided context. Typical CNNs achieve large receptive fields by stacking many convolutional layers, which can be extensively parameter-heavy and computationally expensive for feature maps with high depth.

\begin{figure}[]
    \centerline{\includegraphics[width=\linewidth]{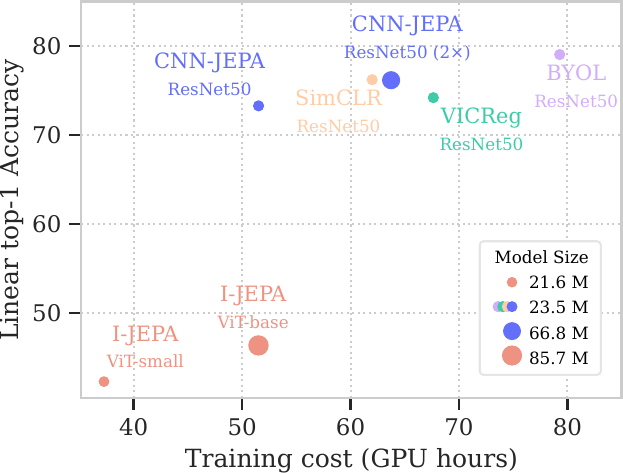}}
    \caption{Comparing CNN-JEPA (our method) to I-JEPA~\cite{SelfSupervisedLearningImages2023a} and common SSL methods based on linear top-1 accuracy, and training cost on ImageNet-100. The area of the markers is proportional to the number of parameters in the model.}
    \label{fig:main_results}
\end{figure}

In this work, we propose a novel self-supervised learning method, which we call CNN-JEPA (Convolutional Neural Network-based Joint Embedding Predictive Architecture), that addresses the challenges of adapting I-JEPA to CNNs. The masked image modeling approach proposed by Tian~et~al.~\cite{DesigningBERTConvolutional2022} uses masked convolutions and masks that are constructed based on the downsampling of the network. We introduce such sparse CNN encoder to I-JEPA to address the challenges of masking. We also introduce a novel, fully convolutional predictor architecture using depthwise separable convolutions that can predict the latent representation of the masked patches with low parameter count and computational cost. Additionally, we improve and simplify the masking strategy of I-JEPA~\cite{SelfSupervisedLearningImages2023a} by predicting masked patches as a single region. 

 Our main contributions are:
\begin{enumerate}
    \item \textbf{A novel self-supervised learning method, CNN-JEPA} that adapts the successful I-JEPA method to Convolutional Neural Networks.
    \item \textbf{Sparse CNN encoder} architecture that can correctly handle masked inputs and produce masked feature maps, building on the masked CNNs proposed by Tian~et~al.~\cite{DesigningBERTConvolutional2022}.
    \item The introduction of a novel, \textbf{fully convolutional predictor} using depthwise separable convolutions.
    \item A \textbf{masking strategy} that follows the downsampling of the network and uses a single target region for the masked prediction task.
\end{enumerate}
    
The source code supporting the findings of this study is available at \url{https://github.com/kaland313/CNN-JEPA}.

\section{Related Work}
We propose to categorize SSL methods that are most relevant to our work into two main groups: instance discriminative and reconstruction-based approaches.
Instance discriminative methods, such as BYOL~\cite{BootstrapYourOwn2020}, SimCLR~\cite{SimpleFrameworkContrastive2020}, and VICReg~\cite{VICRegVarianceInvarianceCovarianceRegularization2022}, learn representations by maximizing similarity between augmented views of the same image. 

Reconstruction-based methods primarily rely on masked image modeling pretext tasks, where parts of the input image are masked, and the network is trained to predict the pixels of the masked regions. These methods were initially proposed for Vision Transformers~\cite{MaskedAutoencodersAre2022}, but were also adapted to Convolutional Neural Networks~\cite{DesigningBERTConvolutional2022}. 

Instance discriminative methods learn image-level features that are invariant to the augmentations used to generate views, and aim to capture high-level semantics of the image.
In contrast, reconstruction-based methods learn more localized features that capture detailed information necessary to solve the reconstruction task; however, their pixel reconstruction objective function might not capture the high-level semantics of the image.
Assran~et~al.~\cite{SelfSupervisedLearningImages2023a} proposed I-JEPA, a method that combines the benefits of both approaches by performing masked image modeling in the latent space, where the network is trained to predict the latent representation of the masked patches. I-JEPA was primarily designed for Vision Transformers, and adapting it to Convolutional Neural Networks has not been explored in the literature to the best of our knowledge.

\section{Method}
\begin{figure*}[]
    \centering
    \includegraphics[width=0.77\linewidth]{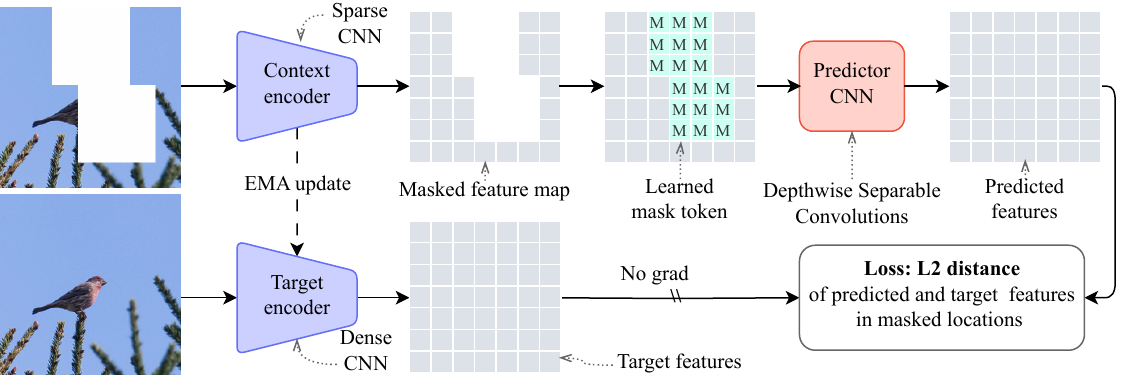}
    \caption{Overview of the CNN-JEPA method. The context and target encoders share a common architecture, with the context encoder using sparse convolutions. The learning objective is to predict the latent representations of masked patches using a predictor, which also trains the context encoder.}
    \label{fig:method_overview}
    \end{figure*}

\subsection{CNN-JEPA Algorithm}

The overview of the proposed CNN-JEPA algorithm is shown on \cref{fig:method_overview}.
The proposed method trains an encoder by masking patches of the input image and predicting the latent representation of the masked patches using a predictor. The target latent representations are obtained by a target encoder whose weights are updated using the exponential moving average of the context encoder's weights.
We introduce a sparse CNN architecture for the context encoder that can correctly handle masked inputs and produce masked feature maps.
For training, we provide the same image to both the context and target encoders, with masking applied only to the context encoder's input. We refer to the masked image and features as the context, based on which the predictor produces the latent representations at the masked locations.
Before feeding the masked latent representation to the predictor, masked locations in the context feature maps are filled with a learnable mask token.
These feature maps are then passed to a predictor, which is a CNN designed to predict the latent representation of the masked patches.

The training objective is to minimize the L2 loss between the predicted and target latent representations at the masked locations. Loss is not computed for non-masked patches.

\subsection{Context and target encoder}

The context and target networks in our proposed method share a common architecture. The context network utilizes sparse convolutions, as introduced by Liu~et~al.~\cite{SparseConvolutionalNeural2015} and later employed by Tian~et~al.~\cite{DesigningBERTConvolutional2022} for masked image modeling. This sparse convolutional architecture is necessary for handling masked inputs and producing masked feature maps.
The target network uses standard convolutions. 

The context network is trained using gradient-based optimization, while the target network's weights are updated using an exponential moving average of the context weights. This update mechanism ensures that the target network provides stable target representations for the loss function. 
In practice, we implement sparse convolutions by selectively masking the outputs of standard convolutional layers. This allows for easily updating the weights of the target network based on the context weights.

\subsection{Predictor}

The predictor in our proposed method operates on feature maps and predicts the latent representation of the masked patches. 
Feature maps typically have high depth (e.g., 2048 for ResNet-50~\cite{DeepResidualLearning2016} and 768 ViT-Base~\cite{ImageWorth16x162020}), but low spatial resolution. As the output of the predictor must have the same dimensionality and shape as the feature maps, its outputs also have such depth and spatial size properties. The number of parameters scales quadratically with the depth for convolutional layers that have the same input and output shape. Hence, for high depth, they require a large number of parameters and are computationally expensive. To address this, we propose to use depthwise separable convolutions, which are more parameter efficient and computationally cheaper than standard convolutions.
With standard convolutions, the number of parameters in the predictor can easily exceed the number of parameters in the encoder. 
Our use of depthwise separable convolutions helps mitigate this issue, ensuring a more balanced distribution of computational resources between the predictor and the encoder.


\subsection{Mask token}
The masked locations in the context feature maps are filled with a learnable mask token, before passing them to the predictor, similarly to how masked image modeling~\cite{MaskedAutoencodersAre2022,DesigningBERTConvolutional2022} and I-JEPA~\cite{SelfSupervisedLearningImages2023a} use a mask token. In our case, the mask token is a parameter vector with the same dimensionality as the depth of the feature maps and is shared across all masked locations and samples of a minibatch.

\subsection{Masking}

In masked image modeling, the most common masking strategy is random masking, where patches are uniformly sampled up to a given masking ratio. I-JEPA~\cite{SelfSupervisedLearningImages2023a} introduces the multi-block masking scheme, where blocks of multiple patches are sampled, patches within blocks are masked, while patches outside the blocks form the context from which the predictor predicts the latent representations. I-JEPA presents an empirical study on random, single, and multi-block masking and shows that the latter is the most effective by a large margin. We adapt the multi-block masking strategy to CNNs. We consider the union of the blocks as the masked region and predict the latent representations in it. The context for the prediction is patches outside any masked patch.

When using masked image modeling with CNNs, the mask size is determined by the downsampling of the network. Typically, ResNets halve the resolution of feature maps in 5 stages (via pooling), resulting in 32 times downsampling. In order to keep the masking consistent with feature map resolution, we use masks with a patch size of $32\times32$ pixels, which corresponds to $1\times1$ spatial dimensions ('pixels') for final feature maps. 

\subsection{Comparison to other methods}
Thanks to its joint embedding predictive architecture, our method learns patch-level features, which present a good trade-off between image-level and pixel-level features. This approach enables the model to capture more localized features compared to instance discriminative algorithms, while also extracting higher-level features than those obtained using masked image modeling methods.

Compared to common instance discriminative SSL methods, one key advantage of our proposed method is its simplicity. It requires minimal handcrafted augmentations, relying only on random resized crop and masking.

Additionally, our approach doesn't require a separate projector network as opposed to most SSL methods, further simplifying the architecture and reducing computational overhead. Many SSL approaches learn features invariant to diverse augmentations, necessitating a projector network after the encoder. 
Our approach learns general features without the need for such a projector network.

\section{Experiments}
\subsection{Encoder and predictor}\label{sec:exp_architecture}
We conduct experiments using a ResNet-50 \cite{DeepResidualLearning2016} encoder for its common use in SSL literature. Sparse convolutions are implemented by zeroing out the outputs of each convolutional layer using a mask appropriately upscaled to the layer's activation dimensions. While this is computationally inefficient, it is simple to implement and can use standard convolutional layers, which are optimized for many GPU architectures. Using sparse convolutional layers would require specialized GPU code, making the implementation of our method more complex and harder to reproduce. 

For the predictor, we use 3 depthwise separable convolutional layers with kernel size 3$\times$3, Batch Normalization, and ReLU activation. The layers of the predictor are arranged sequentially in the following pattern: 3$\times$(DepthwiseConv $\rightarrow$ PointwiseConv $\rightarrow$ BatchNorm $\rightarrow$ ReLU). We conduct ablation studies on the predictor architecture and show that this configuration is the most effective.


\subsection{Classification Probes}
To evaluate the quality of the learned representations, we use linear and k-nearest neighbor (k-NN) classification probes on top of the encoder's frozen features. We train the fully connected layer of the linear probe for 90 epochs and report the best top-1 and top-5 accuracies. For k-NN classification, we use the same features and also report top-1 and top-5 accuracies. For both probes, we use minimal augmentations: deterministic resizing and cropping, and normalization. 

\begin{table}[tbp]
    \begin{threeparttable}
    \centering
    \caption{Linear and k-NN accuracies on ImageNet-1k and ImageNet-100}
    \label{tab:main_results}
    \begin{tabular}{rrcccc}
    \toprule
    {} & {Dataset} & \multicolumn{2}{c}{ImageNet-100} & \multicolumn{2}{c}{ImageNet-1k} \\
    {} & {} & {Linear} & {k-NN} & {Linear} & {k-NN} \\
    {Backbone} & {Algorithm} & {top-1} & {top-1} & {top-1} & {top-1} \\
    \midrule
    \multirow[c]{4}{*}{ResNet-50} & SimCLR~\cite{SimpleFrameworkContrastive2020} & 76.20 & 67.26 & 59.81 & 45.31 \\
     & BYOL~\cite{BootstrapYourOwn2020} & 79.02 & 68.20 & 60.22 & 45.38 \\
     & VICReg~\cite{VICRegVarianceInvarianceCovarianceRegularization2022} & 74.18 & 64.32 & 59.83 & 46.82 \\
     & \textbf{CNN-JEPA [Ours]} & \textbf{73.26} & \textbf{59.22} & \textbf{54.23} & \textbf{35.34} \\
    \midrule
    ViT-Small & \multirow[c]{2}{*}{I-JEPA\cite{SelfSupervisedLearningImages2023a}} & 42.30 & 34.84 & \multicolumn{2}{c}{\multirow{2}{*}{-\tnote{a}}} \\
    ViT-Base & {} & 46.36 & 31.36 &  \multicolumn{2}{c} {} \\
    \bottomrule    
    \end{tabular}
    \begin{tablenotes}
    \item [a]  Assran~et~al.~\cite{SelfSupervisedLearningImages2023a} only published ImageNet-1k results for larger ViT models, which would not provide a fair comparison to ResNet-50. Additionally, we were unable to obtain results for the ViT-Small and ViT-Base models using their official code on our hardware.
    \end{tablenotes}
    \end{threeparttable}
\end{table}

\subsection{Training Details}
We conduct experiments on ImageNet-100 and ImageNet-1k~\cite{ImageNetLargescaleHierarchical2009} datasets. We use a patch size of $32\times32$ pixels and $224\times224$ pixels resolution.
We use the AdamW optimizer with a constant weight decay of 0.01. A cosine learning rate scheduler with warm-up is employed for 10 epochs, reaching a peak learning rate of 0.01. The per-device batch size is 128 with 4 GPUs. 
We update the target encoder's weights using the exponential moving average approach common in SSL methods~\cite{BootstrapYourOwn2020,SelfSupervisedLearningImages2023a}, with the momentum increasing from 0.996 to 1.0 during training.
All pretraining runs use 4 NVIDIA A100 40GB GPUs, 256 GB RAM and 64 CPU cores. Pretraining is conducted for 200 epochs on ImageNet-100 and 100 epochs on ImageNet-1k, taking 13 hours and 70 hours, respectively.



\section{Results}


Our main results are shown in \cref{fig:main_results,tab:main_results}. We compare our method to vision transformers trained with I-JEPA and to common SSL methods that were published for CNNs, such as SimCLR, BYOL, and VICReg. \Cref{fig:main_results} shows that on ImageNet-100 our method outperforms I-JEPA with ViT-Small and ViT-Base by a large margin, and achieves 73.3\% linear top-1 accuracy with a standard ResNet-50 encoder, a result which is competitive with CNN-based SSL methods. Moreover, our method is computationally more efficient than other CNN-based methods due to the simpler, projector-free architecture and the use of only basic augmentations, requiring 17 to 35\% less pretraining hours compared to SimCLR and BYOL respectively. As shown in \cref{tab:main_results}, our method also performs well on ImageNet-1k, achieving a linear top-1 accuracy of 54.23\% in 100 epochs.

\section{Conclusions}
In this paper, we present CNN-JEPA, a novel self-supervised learning method that successfully adapts I-JEPA to convolutional networks. Our method overcomes the challenges of applying masked inputs to CNNs through the use of a sparse CNN encoder and carefully designed masking strategy. To efficiently handle the joint embedding prediction task, our method utilizes a fully convolutional predictor with depthwise separable convolutions. We demonstrated that CNN-JEPA outperforms I-JEPA with ViT architectures on ImageNet-100 and is competitive with other CNN-based SSL methods on both ImageNet-100 and ImageNet-1k, achieving top-1 linear accuracies of 73.26\% and 54.23\% respectively.
CNN-JEPA provides a simpler and more efficient approach to self-supervised learning for CNNs, requiring minimal augmentations and no separate projector network. 

\section*{Acknowledgements}
The work reported in this paper carried out at BME, has been supported by the European Union project RRF-2.3.1-21-2022-00004 within the framework of the Artificial Intelligence National Laboratory.
Project no. TKP2021-NVA-02 and 149336 has been implemented with the support provided by the Ministry of Culture and Innovation of Hungary from the National Research, Development and Innovation Fund, financed under the TKP2021-NVA and MEC\_R\_24 funding schemes respectively.
This research was partly funded by the National Research, Development and Innovation Office of Hungary (FK 142163 grant).
We gratefully acknowledge the support of NVIDIA with the donation of the A100 GPU used for this research and KIFÜ (Governmental Agency for IT Development, Hungary, \url{https://ror.org/01s0v4q65}) for awarding us access to the Komondor HPC facility based in Hungary.

\bibliographystyle{IEEEtran}
\bibliography{IEEEabrv,references}

\end{document}